\documentclass[sigconf]{acmart}
\usepackage[lined,boxed,commentsnumbered,ruled]{algorithm2e}
\usepackage{multirow}
\usepackage{booktabs}
\usepackage{enumitem}
\newtheorem{theorem}{Theorem}
\newtheorem{proposition}[theorem]{Proposition}

\AtBeginDocument{%
  \providecommand\BibTeX{{%
    \normalfont B\kern-0.5em{\scshape i\kern-0.25em b}\kern-0.8em\TeX}}}

\renewcommand\footnotetextcopyrightpermission[1]{} 
\setcopyright{acmcopyright}
\copyrightyear{2023}
\acmYear{2023}
\acmDOI{XXXXXXX.XXXXXXX}

\acmConference[paper]{paper}{2023}{paper}
\acmSubmissionID{180} 

\begin{document}

\title{DiffBFR: Bootstrapping Diffusion Model Towards \\
Blind Face Restoration}
\author{Xinmin Qiu}
\affiliation{%
  \institution{University of Chinese Academy of Sciences}
  \city{Beijing}
  \country{China}
}
\email{qiuxinmin21@mails.ucas.ac.cn}

\author{Congying Han}
\affiliation{%
  \institution{University of Chinese Academy of Sciences}
  \city{Beijing}
  \country{China}}
\email{hancy@ucas.ac.cn}

\author{Zicheng Zhang}
\affiliation{%
  \institution{University of Chinese Academy of Sciences}
  \city{Beijing}
  \country{China}
}
\email{zhangzicheng19@mails.ucas.ac.cn}

\author{Bonan Li}
\authornote{Corresponding author}
\affiliation{%
 \institution{University of Chinese Academy of Sciences}
  \city{Beijing}
  \country{China}}
 \email{libonan@ucas.ac.cn}

\author{Tiande Guo}
\affiliation{%
  \institution{University of Chinese Academy of Sciences}
  \city{Beijing}
  \country{China}}
  \email{tdguo@ucas.ac.cn}

\author{Xuecheng Nie}
\affiliation{%
  \institution{MT Lab, Meitu Inc.}
  \city{Beijing}
  \country{China}}
\email{nxc@meitu.com}



\begin{abstract}
Blind face restoration (BFR) is important while challenging. Prior works prefer to exploit GAN-based frameworks to tackle this task due to the balance of quality and efficiency. However, these methods suffer from poor stability and adaptability to long-tail distribution, failing to simultaneously retain source identity and restore detail. In this paper, we propose to introduce Diffusion Probabilistic Model (DPM) for BFR to tackle the above problem, given its superiority over GAN in aspects of avoiding training collapse and generating long-tail distribution. We name the proposed framework as \textit{\textbf{DiffBFR}}. In particular, DiffBFR utilizes a two-step design, that first restores identity information from low-quality images and then enhances texture details according to the distribution of real faces. This design is implemented with two key components:   
1) Identity Restoration Module (\textbf{IRM}) for preserving the face details in results. Instead of denoising from pure Gaussian random distribution with LQ images as the condition during the reverse process, we propose a novel truncated sampling method which starts from LQ images with part noise added. We theoretically prove that this change shrinks the evidence lower bound of DPM and then restores more original details. With theoretical proof, two cascade conditional DPMs with different input sizes are introduced to strengthen this sampling effect and reduce training difficulty in the high-resolution image generated directly. 2) Texture Enhancement Module (\textbf{TEM}) for polishing the texture of the image. Here an unconditional DPM, a LQ-free model, is introduced to further force the restorations to appear realistic. We theoretically proved that this unconditional DPM trained on pure HQ images contributes to justifying the correct distribution of inference images output from IRM in pixel-level space. Concretely, truncated sampling with fractional time step is utilized to polish pixel-level textures while preserving identity information. Our experiments demonstrated that the proposed DiffBFR achieves significantly superior results to state-of-the-art methods both quantitatively and qualitatively. 
\end{abstract}




\begin{teaserfigure}
\begin{center}
    \includegraphics[width=1.0\linewidth]{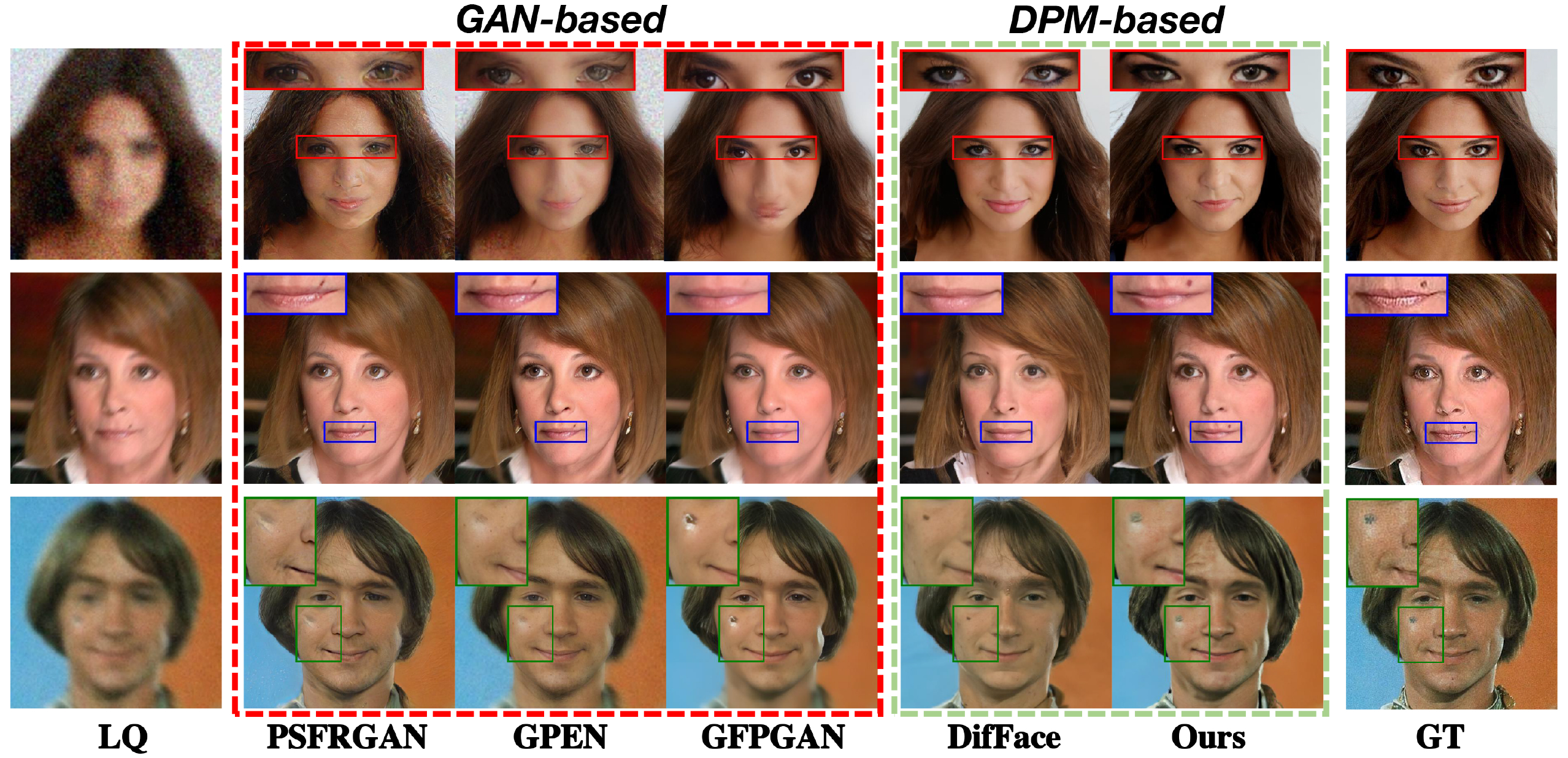}
\end{center}
   \caption{Comparisons of the proposed DiffBFR with state-of-the-art for blind face restoration methods. Left is for low-quality (LQ) images, Middle in the red rectangle for GAN-based results~\cite{PSFRGAN, GPEN, GFPGAN}, Middle in the green rectangle for DPM-based results~\cite{DifFace}, and Right for GroundTruth (GT). We can see DiffBFR achieves better restoration details while maintaining the source identity. Better see in color with 2x zoom.}
\label{RestoredImage0}
\end{teaserfigure}


\title{DiffBFR: Bootstrapping Diffusion Model for \\
Blind Face Restoration}
\maketitle

\begin{figure*}[htbp]
\begin{center}
    \includegraphics[width=1.0\linewidth]{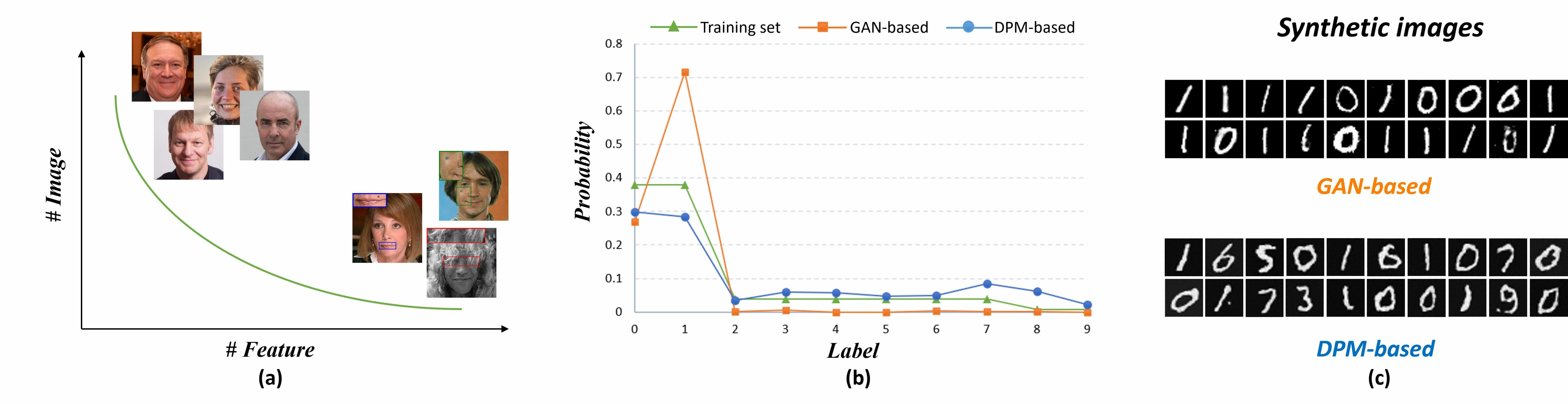}
\end{center}
   \caption{Illustration of long-tail challenge in the BFR task and motivation for our solution based on DPM. (a) The faces sampled from the low-density tail regions in the BFR dataset often comprise novel features, \textit{e.g.}, moles or long fringe, which are hard for existing methods. (b) To address the long-tail challenge, we first investigate the capacities of frequent generative models on a toy long-tail MNIST dataset with $28 \times 28$ resolution, where images with labels other than 0 and 1 are partially dropped. (c) The random syntheses combined with statistical data in (b) demonstrate that the GAN-based model fails to synthesize high-fidelity datapoints from low-density regions , while the DPM-based model shows promising results in addressing this problem.}
\label{mnist_frame}
\end{figure*}

\section{Introduction}
Blind face restoration (BFR)~\cite{GFPGAN, GPEN, VQFR} aims to recover high-quality (HQ) face images from the low-quality (LQ) ones. It is an important task in Computer Vision and Graphics communities and is widely applied in various scenarios, \emph{e.g.}, monitoring image restoration, old photos restoration and face image super-resolution, etc. However, this task is very challenging due to the non-deterministic degradation that harms the image quality, such as blurring, noising, down-sampling and compression artifacts. 



Previous works for BFR typically rely on Generative Adversarial Networks (GAN). They mainly focus on designing various face-specific priors to tackle the problem, including generative priors~\cite{GFPGAN, GPEN}, reference priors~\cite{DFDNet, GFRNet} and geometric priors~\cite{yu2018face, PSFRGAN}. Although achieving state-of-the-arts, these methods still encounter difficulties in restoring fine-grained facial details while achieving realistic texture, as illustrated in Figure.~\ref{RestoredImage0} and Figure.~\ref{RestoredImage2}. This can be ascribed to the fact that these methods generally need to project degraded images into the latent space of a pre-trained GAN, \textit{e.g.}, StyleGAN~\cite{styleGAN}, while the limited capacity of GANs makes the projection difficult to exactly retain the content details of given images. In addition, GAN-based models face the problem of "training collapse" as the objective function is a min-max function and optimization is difficult. Moreover, due to the poor adaptability of GAN-based methods to long-tail distributed dataset~\cite{long_tail}, restored faces derived by GAN are prone to change person identities, and they are hard to achieve the balance between image restoration quality and character fidelity maintenance. We conduct experiments on a toy long-tail distribution MNIST to verify this problem and results are shown in Figure.~\ref{mnist_frame}, which proves that 
the GAN-based model~\cite{gan} fails to cover low-density regions and can not generate local details as the tail feature in the face dataset. It is critical to address these issues for advancing practical applications of BFR in the real world.

To achieve the above goal, we propose to bootstrap diffusion models for blind face restoration in this paper, further pushing forward the frontier of this task. Our main motivation is the superiority of diffusion models over GANs in aspects of avoiding training collapse and generating long-tail distribution. We name our method as \textit{\textbf{DiffBFR}}. In particular,
DiffBFR exploits 
Diffusion Probabilistic Models (DPM) for enhancing the face-specific prior, considering its great power to produce HQ images in the wild range of distribution. For deriving accurate restoration of LQ faces, DiffBFR utilizes a novel component capturing and texture polishing strategy. Specifically, for component capturing, DiffBRF proposes to denoise from the LQ image and a diffused version of it, which shrinks the evidence lower bound of DPM with theoretical proof and further helps to maintain more details. For texture polishing, DiffBFR relies on the analysis of the similarity of noise space and then exploits rich priors from pure HQ images, which helps to 
synthesize factual images with natural texture. In this way, DiffBFR completes blind face restoration in two steps: first to restore the content information from LQ images and then to enhance the texture of images, thus producing reliable restoration results.

In particular, DiffBFR is composed of two core modules: (1) Identity Restoration Module (IRM). IRM aims to capture facial information in LQ images. Here, IRM begins with a conditional DPM at low resolution, followed by one conditional super-resolution DPM that upsamples the image. Compared with direct training on large-resolution images, the model can converge faster and obtain better results. During the sampling phase, IRM performs a full reverse diffusion process with low-resolution DPM. For super-resolution DPM, IRM proposes a novel truncated sampling strategy, that is, denoising from intermediate diffused variables, to efficiently preserve more details in results. (2) Texture Enhancement Module (TEM). TEM is designed to polish the texture of images. Specifically, we train an unconditional DPM with pure HQ images and perform denoising starting from the noisy version of the output from IRM. As a result, texture information with a high degree of naturalness is recorded without any impact from LQ images. The use of TEM sharpens the edge structure and further forces the restorations to appear realistic.

From the perspective of both experimental exploration and theoretical derivation, we show that the proposed DiffBRF effectively deploys diffusion models for solving the blind face restoration problem, which not only reduces the training difficulty and training time of the whole model, but also provides less degradation serious conditional input for Truncated Sampling Module. 
Our contributions can be summarized into three folds: \\(1) To the best of our knowledge, we are the first to propose the application of pure diffusion models to the task of blind face restoration, motivated by its superiority over GANs on avoiding training collapse and generating long-tail distribution. \\(2) We present two novel modules in DiffBFR: Identity Restoration Module (IRM) and Texture Enhancement Module (TEM), which effectively restores high-fidelity facial details while maintaining person identities. Additionally, we also theoretically proved that they can yield better recovery results in the inference process. \\(3) Through extensive experiments, we demonstrate that DiffBFR sets new state-of-the-arts on multiple benchmarks for the blind face restoration task.


\section{Related Work}
\paragraph{\textbf{Image Restoration}}
Image Restoration usually includes super-resolution~\cite{imageSR}, denosing~\cite{denoiser}, deblurring~\cite{deblurgan}, compression removal~\cite{compression} and their random combination and so on, which is classical research in the field of computer vision. In the past, most image restoration problems were based on the image degradation model known to give corresponding restoration methods, such as DnCNNs~\cite{denoiser}, DeblurGAN~\cite{deblurgan}, etc. However, in the real world, the degradation causes of LQ images that need to be restored are mostly unknown. How to restore images whose degradation ways are unknown is an important challenge in this research field in recent years.

\paragraph{\textbf{Blind Face Restoration}}
Blind Face Restoration~\cite{GFPGAN, GPEN, VQFR} is an important branch in the field of Image Restoration. Its task objective is to restore low-quality (LQ) face images into high-quality (HQ) ones on the premise that degradation models and parameters are completely unknown. In recent years, great breakthroughs have been made in the BFR task, such as the method based on geometric prior of face~\cite{yu2018face, PSFRGAN}, the method based on reference prior~\cite{DFDNet, GFRNet}, and so on. GFPGAN~\cite{GFPGAN} and GPEN~\cite{GPEN} embed face prior information using a GAN-based generation model which uses an encoding-decoding frame. PSFRGAN~\cite{PSFRGAN} combined the structural features of face segmentation and proposed a GAN-based progressive restoration network. VQFR~\cite{VQFR} combines the classical dictionary-based method with the recent vector quantization (VQ) technology.

\paragraph{\textbf{Diffusion Probability Models}}
In the past few years, GAN-based generative models have been almost the mainstream, and after the proposal of Denoising Diffusion Probabilistic Models (DDPM)~\cite{DDPM, IprovedDDPM} and Denoising Diffusion Implicit Models (DDIM)~\cite{DDIM}, the generative model based on diffusion models~\cite{diffusion} has become a breakthrough in the field of computer vision with its excellent image generation quality advantage~\cite{dreamfusion,ei,styo}. GAN-based model training is prone to collapse, which is avoided by the diffusion model method. This diffusion-based approach has attracted considerable attention from computer vision and natural language processing to graphic analysis. CARD~\cite{CARD} proposed a classification and regression diffusion model, combining a conditional generation model based on denoising diffusion and a pre-trained conditional mean estimator to predict the data distribution under a given condition. Inspired by CLIP~\cite{CLIP}, GLIDE~\cite{GLIDE} explored real image synthesis with text conditions and found that diffusion models with class-free guidance produced high-quality (HQ) images that included a wide range of learned knowledge. With the help of a variational auto-encoder framework, the diffusion model of latent space training is established by LSGM~\cite{LSGM}. SegDiff~\cite{SegDiff} extends the diffusion model to perform image-level segmentation by summarizing feature maps from the diffusion probabilistic encoder and the image feature encoder.

\begin{figure*}[htbp]
\begin{center}
    \includegraphics[width=1\linewidth]{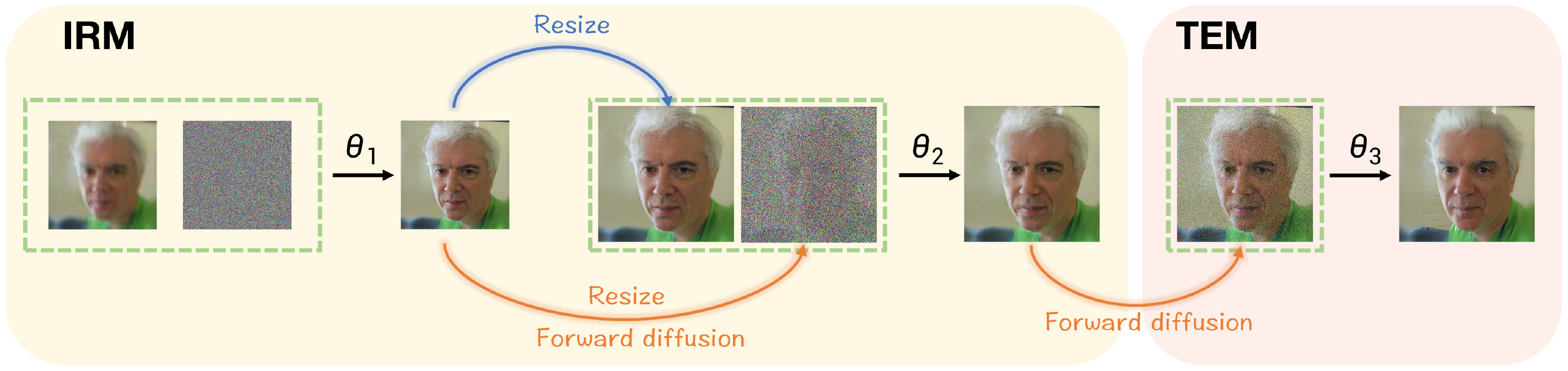}
\end{center}
   \caption{Sampling process of the proposed \textit{DiffBFR} for blind face restoration task. In essence, DiffBFR is a cascaded diffusion model: Given a LQ face, an Identity Restoration Module (\textit{IRM}) enriches the facial details at both low- and high-resolution successively, and a Texture Enhancement Module (\textit{TEM}) further polishes the realistic texture of the image to predict the HQ face. The DPM-based design of DiffBFR confers advance in performance verified by both theoretical and practical evidence. } 
\label{framework}
\end{figure*}

\section{Preliminaries}
In this section, we briefly introduce fundamental notations and definitions to facilitate comprehension~\cite{DDPM, SR3, CascadedDM} of our proposal. 

\paragraph{\textbf{Denoising Diffusion Probability Models}}
 DDPM~\cite{DDPM} establishes a relationship between a complex distribution $p(y)$ and the Gaussian distribution $N(0, I)$ using forward and reverse Markov chains. Following the convention, we denote $y$ as $y_0$, and the forward process generates latent variables $y_1,\dots,y_T$ through 
 \begin{equation}
		\begin{aligned}
			q(y_t|y_{t-1})&=N(y_t;\sqrt{\alpha_t}y_{t-1}, (1-\alpha_t)I). 
		\end{aligned}
\end{equation}
where $\{\alpha_t\}_{t=1}^T$ is a fixed variance schedule rather than learned parameters. The forward process holds the property
\begin{equation}\label{eq:forward}
		\begin{aligned}
			q(y_t|y_0)&=N(y_t;\sqrt{\gamma_t}y_0, (1-\gamma_t)I),
		\end{aligned}
\end{equation}
where $\gamma_t=\prod_{i=1}^{t}\alpha_i$. The reverse process starts from $y_T$ to sample the real data $y_0$ sequentially through
\begin{equation}\label{eq: reverse}
    \begin{aligned}
        p_{\theta}(y_{t-1}|y_t)=N(y_{t-1};\mu_{\theta}(y_t,t), \Sigma_{\theta}(y_t,t)),
    \end{aligned}
\end{equation}
where $\mu_{\theta}$ is a parameterized function to be trained for maximizing evidence lower bound (ELBO) of $p(y_0)$, and $\Sigma_{\theta}=\sigma_t^2 I$ where $\sigma_t$ is usually a pre-defined constant related to the variance schedule. Further, by decomposing $\mu_\theta$ into a linear combination of $x_t$ and the noise approximator $\epsilon_\theta$, the generative process can be expressed in another form:
\begin{equation}\label{eq: creverse}
    \begin{aligned}
        x_{t-1}=\frac{1}{\sqrt{\alpha_t}}(x_t-\frac{1-\alpha_t}{\sqrt{1-\gamma_t}}\epsilon_\theta(x_t,t))+\sigma_t\epsilon,
    \end{aligned}
\end{equation}
where $\epsilon\sim N(0,I)$, which suggests that each generation step is stochastic. Similarly, a conditional distribution $p(y_0|x)$ can be approximated by the diffusion process:
\begin{equation}\label{eq: creverse_cond}
    \begin{aligned}
        p_{\theta}(y_{t-1}|y_t, x)=N(y_{t-1};\mu_{\theta}(y_t,t,x), \Sigma_{\theta}(y_t,t)).
    \end{aligned}
\end{equation}

\paragraph{\textbf{Cascaded Diffusion Model}}
Cascaded diffusion model~\cite{CascadedDM} (CDM) is an effective method to scale a diffusion model to high-dimension distribution. Specifically, for a high-resolution image $y_{0}$, an extra latent variable (\textit{e.g.}, down-sampled image) $z_0$ that is easier to learn than $y_0$ is introduced, thus we can reformulate the generative process of $y_0$ as
\begin{equation}\label{eq:cdm}
    p(y_0) = \int p(y_0|z_0) p(z_0) dz_0,
\end{equation}
 which corresponds to a two-stage cascaded model. In this way, CDM first learns a diffusion model as Eq.\eqref{eq: reverse} for low-resolution image $z_{0}$, then learns a conditional diffusion model as Eq.\eqref{eq: creverse} to sample $y_{0}$ from $p(y_0|z_0)$. In practice, the cascaded process can be divided into multiple stages by inserting additional latent variables.

\section{Methodology}
In this section, we present DiffBFR, a diffusion probability model designed to address the BFR task. As depicted in Figure.~\ref{framework}, DiffBFR primarily comprises two fundamental modules: the Identity Restoration Module (IRM) and the Texture Enhancement Module (TEM). IRM learns to straightforwardly enhance facial identity details at both low- and high-resolution levels,  achieving superior identity preservation. TEM further refines the realistic texture of the image with a DPM-based facial texture prior, enabling the prediction of HQ face images. 

Unlike previous methods that project LQ images into the vectorized and compact latent space of pre-trained GANs, potentially resulting in texture and identity information loss, the proposed DiffBFR offers a more intuitive solution for enhancing image details in a non-compressed and expressive latent space, while preserving facial details.  In the following, we begin with a comprehensive analysis of the BFR task and DiffBFR's mechanism (Sec.~\ref{sec: overall}), then delve into the technical details of the IRM (Sec.~\ref{sec:irm}) and TEM (Sec.~\ref{sec:tem}) components to illustrate the advantages of our proposal.
\subsection{Long-tail challenge in BFR task}\label{sec: overall}
We approach the BFR problem from the lens of conditional generation: Given a dataset consisting of various LQ-HQ image pairs, we aim to learn a conditional distribution $p(y_0|x_0)$, in which $x_0$ and $y_0$ denote the LQ and HQ variables, respectively. Empirically, the data in BFR dataset are typically scattered across a high-dimensional space with a {long-tail} distribution~\cite{long_tail}: The head region of distribution only comprises a limited number of normal cases, whereas the long-tail region consists of numerous hard cases, \textit{e.g.}, grayscale image and face with moles.
Unlike in the classification task~\cite{long_tail2}, the low-level feature appearing on the tail part refers to attributes that less influence the identity,  but is important for visual effects.

We state that \textit{learning such a long-tail distribution $p(y_0|x_0)$ poses a significant challenge for existing BFR methods.}
As evidenced by Figure.~\ref{RestoredImage0}, previous GAN-based works cannot well tackle the samples residing in both head and long-tail regions, resulting in obvious over-smoothing texture as well as distorted content compared to the GroundTruth. DiffFace~\cite{DifFace}, a concurrent DPM-based method, also encounters similar issues. Given the practical limitations of existing methods, it is significant to tackle the challenge of advancing the frontier of BFR.  

\subsection{DiffBFR: Diffusion model for BFR task}
In this part, we explore overcoming the challenge via a reasonable design to well approximate the long-tail distribution $p(y_0|x_0)$. Our proposal named DiffBFR, a DPM-based model for BFR task, has two main advantages: (\textbf{i}) Clear theoretical strengths and interpretability.  (\textbf{ii}) Concise and easy to train in practice. As presently there are two mainstream generative models to learn a distribution, \textit{i.e.}, GAN and DPM, we first answer the following question to strengthen the rationale of choosing DPM rather than GAN as the base model:
\paragraph{\textbf{DPM or GAN, which one is the most promising to solve the long-tail challenge?}} As shown in Figure.~\ref{mnist_frame}, we provide a toy dataset and models to explain that DPM would be the solution. At first, we construct a long-tail MNIST dataset with $28 \times 28$ resolution. Compared with the vanilla MNIST~\cite{deng2012mnist}, we partially discard some samples, such that the images with labels 0 and 1 have a higher density and the others have a lower density. Then, we train toy DDPM~\cite{DDPM} and GAN~\cite{gan} on the long-tail dataset with $28 \times 28$ resolution, in which generators have similar numbers ($\sim$1.5 million) of parameters. After that, we count the labels of random samples from trained DPM and GAN. The results demonstrate that the DPM is promising to align the long-tail distribution reasonably, whereas GAN tends to fit the head region with high density, resulting in a very low probability of label generation in the long-tail, and even a few categories are barely generated anymore. Therefore, we design DiffBFR as a DPM-based model to better solve the challenge.
\paragraph{\textbf{Cascaded structure in DiffBFR}} 
Although DDPM performs better in the toy long-tail dataset, in practice the large size ($\geq$$512^2$ pixels) and scale ($>$50k) of BFR datasets make it non-trivial to directly apply it to the BFR task.  We find that a proper design of cascaded structure can not only enhance training stability~\cite{CascadedDM}, but also improve the quality of restoration. Specifically, DiffBFR is based on the reformulation: 
\begin{subequations}\label{eq:diffbfr}
\begin{align}
        p(y_0|x_0) & = \int p(y'_0,x'_0|x_0)p(y_0|y'_0, x'_0,x_0) dx'_0dy'_0, \tag{\ref{eq:diffbfr}{a}}  \label{eq:diffbfr a}\\
                & \approx \int \underbrace{p(x'_0|x_0) p(y'_0|x'_0)}_{IRM} \underbrace{p(y_0|y'_{0})}_{TEM}dx'_0dy'_0. \tag{\ref{eq:diffbfr}{b}}  \label{eq:diffbfr b}
\end{align}
\end{subequations}
Herein, we introduce two new intermediate variables $x'_0$ and $y'_0$ with the same shape of $x_0$ and $y_0$, respectively. The formulation of DiffBFR first follows and inherits the advantages of CDM in training speed and stability, where each conditional and unconditional distribution in Eq.\eqref{eq:diffbfr b} can be approximated by SR3~\cite{SR3} or DDPM~\cite{DDPM}. Moreover, beyond just outperforming in model training, we note that each module in DiffBFR with its specific design will enhance the prediction for the BFR task. The first one is called the Identity Restoration Module where (IRM) upsamples the LQ image $x_0$ to gradually arrive at the resolution of $y_0$ while enriching the facial details, and the second one called TEM exploits the diffusion-based facial prior to further refining the texture details. Both IRM and TEM are equipped with \textit{truncated sampling strategies}, alleviating the unfaithful results due to excessive noise in Eq.\eqref{eq:forward}. The remaining part elaborates on technical details.

\subsection{Identity Restoration Module}\label{sec:irm}
Given each training LQ-HQ pair $(x_0, y_0)$, the IRM learns the cascaded conditional distribution to map LQ image $x_0$ into the high-resolution image with two steps. The first stage first enriches the facial details at a low resolution as same as $x_0$, where a DDPM is trained with the objective 
\begin{equation}
    \mathop{\min}_{\theta_1} \mathbb{E}_{x_0, \epsilon \sim N(0,I), t\sim \text{Uniform(1, T)}} \left\| \epsilon - \epsilon_{\theta_1}(x'_t, t, x_0) \right\|^2_2.
\end{equation}
 $x'_0 $ is the low-resolution GroundTruth downsampled from $y_0$ with a scale factor $r$, \textit{i.e.}, $x'_0 = [y_0]{\downarrow}_{r}$, and $x'_t$ is the noisy image of $x'_0$ sampled from  Eq.\eqref{eq:forward}. 
We denote the sample from learned distribution as $\tilde{x}'_{0}$. Then a DDPM is trained with the following objective
\begin{equation}
    \mathop{\min}_{\theta_2} \mathbb{E}_{x_0, \epsilon \sim N(0,I), t\sim \text{Uniform(1, T)}} \left\| \epsilon - \epsilon_{\theta_2}(y'_t, t, \tilde{x}'_0) \right\|^2_2.
\end{equation}
We provide more training details of $\epsilon_{\theta_1}$ and $\epsilon_{\theta_2}$ in Experiments (Sec.~\ref{sec4}) and Supplementary Materials.

\paragraph{\textbf{Truncated sampling}}
The sampling strategy in the reverse process~\cite{SDEdit} based on Eq.\eqref{eq: creverse} has a crucial impact on the quality of results. For the BFR task, we find the way starting with $y'_T \sim N(0, I)$ to sample from  $p_{\theta}(y'_{t-1}|y'_t, \tilde{x}'_0)$ subsequently cannot exploit the full potential of the trained DDPMs, where the final result $y'_0$ are probably unfaithful to $\tilde{x}'_0$ in terms of identity. Therefore, we propose a truncated sampling strategy in the conditional frame to improve it. 
The reverse process will be conditioned on $y_{N_1}$, where the truncated time $N_1 < T$. In the following proposition, we provide a theoretical analysis of the advantage of truncated sampling compared with vanilla sampling.

\begin{proposition}\label{prop:1}
Given a LQ image $x_0$ and HQ image $y_0$, 
we denote the evidence lower bound (ELBO) of vanilla diffusion, and diffusion with truncated sampling as $L_{DDPM}$ and $L_{IRM}$, respectively. Then, we have
    \begin{equation}
		\begin{aligned}
			L_{DDPM}\leq L_{IRM}.
		\end{aligned}
    \label{prop1}
    \end{equation}
\end{proposition}

 Proposition \ref{prop:1} shows that for conditional DDPM, the change of truncated sampling can shrink the ELBO of the model. Furthermore, it can be proved that the higher the quality of the condition input $\tilde{x}'_0$, the closer it is to $y_0$, the more accurate the restored image will be. This explains why we need to restore low-resolution images first in IRM. In a nutshell, we design IRM as follows: the restoration preprocess on low-resolution images provides an input, so that the conditional DPM on high-resolution ones can generate higher-quality images with these effective sampling changes. 

\begin{figure*}[htbp]
\setlength{\abovecaptionskip}{0.cm}
\setlength{\belowcaptionskip}{-0.cm}
\begin{center}
    \includegraphics[width=0.95\linewidth]{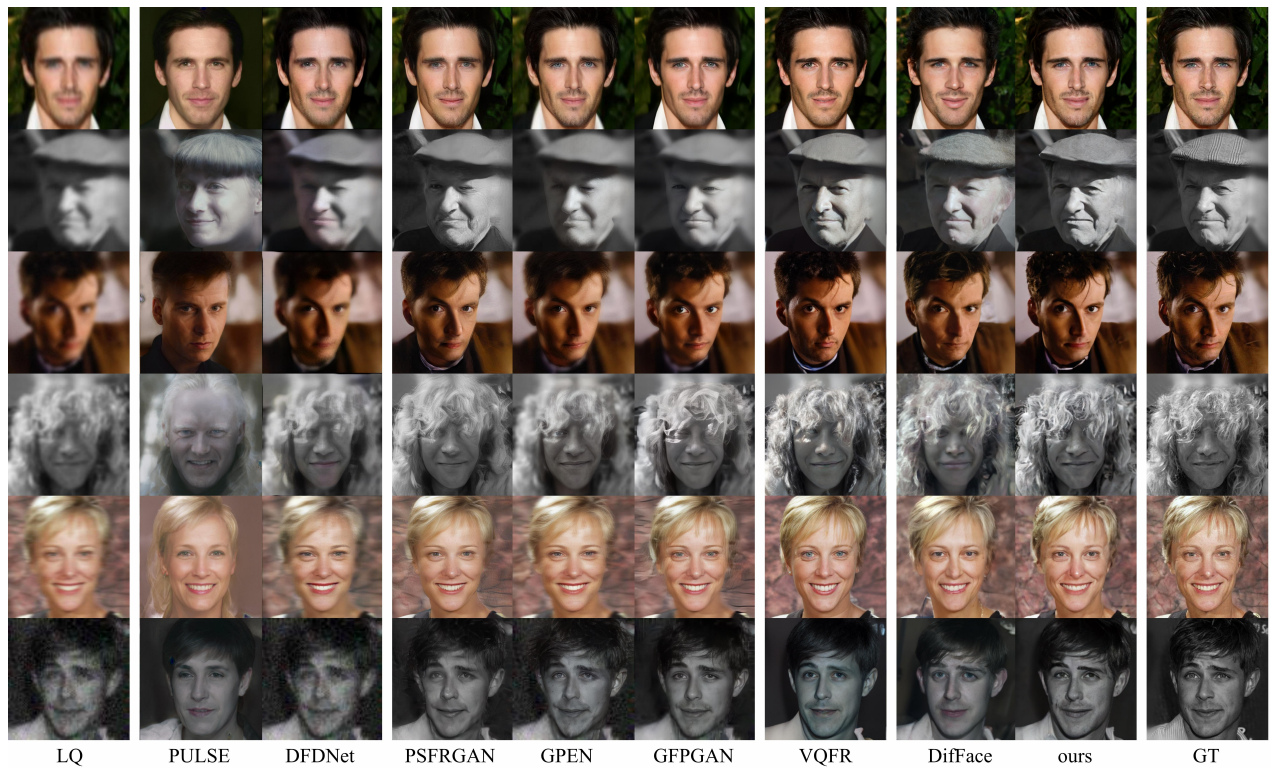}
\end{center}
   \caption{Qualitative comparisons on the \textbf{CelebA-Test} for blind face restoration and from left to right: low-quality image, PULSE~\cite{PULSE}, PSFRGAN~\cite{PSFRGAN}, GPEN~\cite{GPEN}, GFPGAN~\cite{GFPGAN}, VQFR~\cite{VQFR}, DifFace~\cite{DifFace}, our DiffBFR and GroundTruth. Our DiffBFR performs well in both detail complement and hue preservation. \textbf{Zoom in for best view}.}
\label{RestoredImage2}
\end{figure*}

\begin{table*}
\setlength{\abovecaptionskip}{0.cm}
\setlength{\belowcaptionskip}{-0.cm}
\centering
\setlength\tabcolsep{4.5pt}
\caption{Quatitative comparison on \textbf{CelebA-Test} with 3000 images randomly for blind face restoration. \textcolor{red}{Red}, \underline{underline} and \textcolor{blue}{blue} indicate the best, the second best and the third best performance.}
\label{result}
\begin{tabular}{c|c|ccccccc|c}
\bottomrule
\multirow{2}*{Metrics} & \multirow{2}*{Input(LQ)} & \multicolumn{8}{|c}{Methods}\\
\cline{3-10}
~ & ~ & DFDNet\cite{DFDNet} & PULSE\cite{PULSE} & GPEN\cite{GPEN} & GFPGAN\cite{GFPGAN} & PSFRGAN\cite{PSFRGAN} & VQFR\cite{VQFR} & DifFace\cite{DifFace} & \textbf{DiffBFR}(ours)\\
\hline
SSIM$\uparrow$& 0.6460&	0.6444&	0.6102&	\underline{0.6777}&	\textcolor{red}{0.6827}&	0.6213&	0.6382&	0.6494&	\textcolor{blue}{0.6553}\\
 PSNR$\uparrow$& 24.921&	23.300&	21.619&	\textcolor{red}{25.423}&	\underline{25.401}&	24.596&	23.568&	24.055&	\textcolor{blue}{24.748}\\
\hline
 FID$\downarrow$& 93.564&	39.649&	45.940&	22.507&	20.676&	26.050&	\underline{17.862}&	\textcolor{blue}{19.653}&	\textcolor{red}{16.490}\\
 NIQE$\downarrow$ & 9.1407&	6.4226&	7.3754&	6.7775&	6.7324&	\underline{5.6114}&	\textcolor{blue}{5.9606}&	6.1638&	\textcolor{red}{5.5990}\\
 LPIPS$\downarrow$ & 0.5953&	0.3901&	0.4209&	0.2956&	\textcolor{blue}{0.2823}&	0.3101&	\underline{0.2616}&	0.3052&	\textcolor{red}{0.2535}\\
\toprule
\end{tabular}
\end{table*}

\subsection{Texture Enhancement Module}\label{sec:tem}
Despite the delicate facial details can be well restored via IRM, we experimentally find that the results usually retina some weird texture, such as the edge on the corners of the eyes, teeth and other facial features, which are obvious to impede the visual effect. We conjecture that this unnatural texture may result from the excessive restoration of IRM. 
In the end, we find imposing a diffuse-based facial prior to restored faces from IRM can greatly remove texture weakness. We train an \textit{unconditional} DDPM with the objective
\begin{equation}
    \mathop{\min}_{\theta_3} \mathbb{E}_{y_0, \epsilon \sim N(0,I), t\sim \text{Uniform(1, T)}} \left\| \epsilon - \epsilon_{\theta_3}(y_t,t) \right\|^2_2.
\end{equation}
In this way, the sampling starts from $y_{N_2}\sim q(y_{N_2}|y'_{0})$ that sampled from Eq.\eqref{eq:forward} indeed formulate $p(y_0|y'_{0})$ to enhance the texture details of restored faces, which names TEM. 

Moreover, by cooperating with Fréchet Inception Distance in theory, we prove that TEM can effectively correct the distribution of the restoration images.
\begin{proposition}\label{prop:2}
Assume that the LQ image input is $x$, the HQ image is $y$, and the inference image is $y'$. It can be proved that the FID of the resulting image distribution after TEM is lower than that before TEM. We have
    \begin{equation}
		\begin{aligned}
			FID(x,y)>FID(y',y).
		\end{aligned}
    \end{equation} 
 \end{proposition}
 Proposition \ref{prop:2} is precisely proving that the FID of the inference that images distribution after TEM is lower than that before TEM, and the obtained inference images have a more similar distribution than HQ images on the whole.

\section{Experiments}\label{sec4}

In this section, we introduce the training dataset, testing dataset in Sec.~\ref{sec::datasets} and specific experimental results comparison in Sec.~\ref{sec::comparison}. We perform ablation studies to demonstrate the effectiveness of the proposed IRM and TEM in Sec.~\ref{sec:ablation}.

\begin{figure*}[htbp]
\setlength{\abovecaptionskip}{0.cm}
\setlength{\belowcaptionskip}{-0.cm}
\begin{center}
    \includegraphics[width=0.8\linewidth]{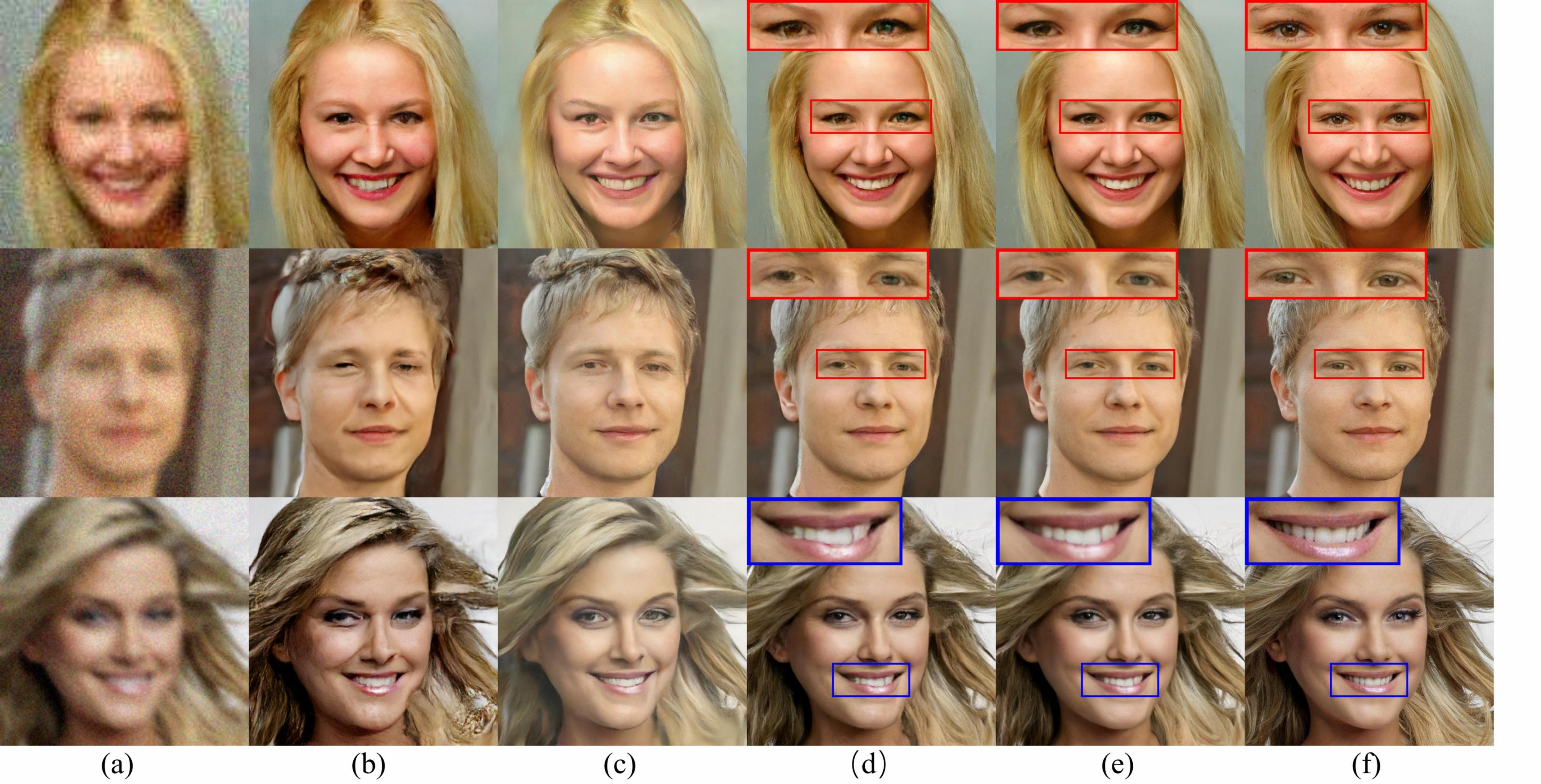}
\end{center}
   \caption{Qualitative comparisons on \textbf{CelebA-Test} for blind face restoration in ablation results. (a) LQ images, (b) IRM-s: 1-stage DPM without cascade, (c) IRM-c: 2-stage DPM with cascade, (d) IRM-c-t: 2-stage DPM which is added truncated sampling module in the second stage, namely IRM(-2), (e) TEM-w: 3-stage DPM which is added unconditional justify module in TEM, (f) GT images. \textbf{Zoom in for best view}.}
\label{AblationResultFigure}
\end{figure*}

\subsection{Datasets}\label{sec::datasets}
\paragraph{\textbf{Training Datasets}}
We choose FFHQ~\cite{styleGAN} as the training dataset, which contains 70,000 high-quality PNG format face images with $1024\times1024$ resolution. In this experiment, we resize all images to $512\times 512$ to train face restoration at this resolution.

Since our DiffBFR is supervised training, the corresponding LQ-HQ image pairs are required. We use generated random degradation model to simulate LQ images in the real world. Its generation formula~\cite{degradation2, GFPGAN} is shown in Eq.\eqref{formula_degradation}, where $y$ is the HQ image, $k_{\sigma}$ is the Gaussian blur kernel, $r$ represents the down-sampling scale factor, and $q$ represents the JPEG compression of the image with quality factor $q$. In order to keep the experimental results directly comparable, the parameters $\sigma$, $r$, $\delta$, $q$ are randomly sampled from \{0.1: 10\}, \{0.8: 8\}, \{0: 20\}, \{60: 100\}, respectively, to align with the experimental environment of recent methods for BFR task. We also add gray color probability during the training process for color adaptation and augment data with the horizontal flip.
\begin{equation}
    \begin{aligned}
        x=[(y\otimes k_{\sigma})\downarrow_{r}+n_{\delta}]_{\text{JPEG}_{q}}
    \end{aligned}
    \label{formula_degradation}
\end{equation}
\paragraph{\textbf{Testing Datasets}}
We choose CelebA-Test as the testing dataset, which contains 3,000 HQ images randomly sampled from CelebA-HQ~\cite{Celeba} with the resolution of $512\times 512$. Similarly, the corresponding random LQ images are generated for evaluation by using the degradation model in Eq.\eqref{formula_degradation} and the same set of parameters used in the training dataset. Our method and other state-of-the-art methods are tested on the same CelebA-Test dataset to observe their quantitative comparisons and qualitative comparisons.

\subsection{Comparisons with State-of-the-art Methods}\label{sec::comparison}

\paragraph{\textbf{Comparison Methods}}
During the experiments, we noted the concurrent work, DifFace~\cite{DifFace}, which is also included in the comparisons.
We compare DiffBFR with seven recent BFR methods, including DFDNet~\cite{DFDNet}, PULSE~\cite{PULSE}, PSFRGAN~\cite{PSFRGAN}, GFPGAN~\cite{GFPGAN}, GPEN~\cite{GPEN}, VQFR~\cite{VQFR}, and DifFace.
\paragraph{\textbf{Metrics}}
We quantitatively compare the differences between our method and state-of-the-art methods using five widely-used metrics, including SSIM~\cite{SSIM}, PSNR, FID~\cite{FID}, NIQE~\cite{NIQE}, and LPIPS~\cite{LPIPS}. Among them, NIQE is a no-reference metric. SSIM and PSNR are pixel-wise similarity measures, while FID, NIQE and LPIPS are perceptual measures. 

\paragraph{\textbf{Quantitative Results}}
As shown in Table.~\ref{result}, the comparison results on the CelebA-Test are summarized and our method shows better results in quantitative results. DiffBFR achieves the best FID, NIQE and LPIPS scores, indicating that our restoring results are close to the real face image distribution and the natural image distribution and maintain the perceptual approximation to GroundTruth. However, the pixel-wise metrics SSIM and PSNR are not highly correlated with the subjective evaluation of human observation. DiffBFR only maintains a relatively similar degree with recent state-of-the-art methods in these two metrics to achieve the basic goal of the restoration task, which is not good at these two measures. 
\begin{table}
\setlength{\abovecaptionskip}{0.cm}
\setlength{\belowcaptionskip}{-0.cm}
    \begin{center}
    \setlength\tabcolsep{7.0pt}
    \caption{Ablation study results on \textbf{CelebA-Test} for blind face restoration. \underline{\emph{IRM-s}}: use 1-stage DPM in IRM; \underline{\emph{IRM-c}}: use 2-stage cascade DPM in IRM; \underline{\emph{IRM-c-t}}: change the sampling process in the second stage in truncated sampling; \underline{\emph{TEM-w}}: add the advanced unconditional DDPM in TEM.}
    \label{ablationResult}
    \begin{tabular}{l|cc|ccc}
    \bottomrule
    Method & SSIM$\uparrow$ & PSNR$\uparrow$ & FID$\downarrow$ & NIQE$\downarrow$ & LPIPS$\downarrow$ \\
    \hline
     IRM-s & 0.5266&	22.8438&	31.3126&	6.3403&	0.4873 \\
     IRM-c & 0.5879&	21.5117&	24.2364&	5.8538&	0.3378\\
     IRM-c-t & 0.6494&	24.727&	19.6023&	\textcolor{red}{5.4831}&	0.2546 \\
     TEM-w & \textcolor{red}{0.6553}&	\textcolor{red}{24.7485}&	\textcolor{red}{16.4902}&	5.5990&	\textcolor{red}{0.2535} \\
    \toprule
    \end{tabular}
    \end{center}
    \vspace{-6mm}
\end{table}
\paragraph{\textbf{Qualitative Results}}
Figure.~\ref{RestoredImage2} shows the restoration effect comparison of color images and gray images. Obviously, our method can see the restoration ability of the face in the visual image. Due to the inclusion of the conditional module, DiffBFR maintains quite good results in fidelity. From the figure, we can see that in LQ images with serious degradation, DiffBFR is able to obtain inference images without blurring and significant noise residual. Additionally, for color images and gray images, DiffBFR can maintain the same color intensity as the GroundTruth as much as possible, which plays an important role in the restoration of light and shadow effects in image restoration. From Figure.~\ref{RestoredImage2}, we can see that PULSE~\cite{PULSE} changes the identity during the restoration process, and the restored face of the severely degraded image is not the same person from the human point of view. DFDNet~\cite{DFDNet} has a limited ability to restore the face structure, and many details keep the blurred part in the LQ image, which can not supplement the clearer HQ image. PSFRGAN~\cite{PSFRGAN}, GPEN~\cite{GPEN} and GFPGAN~\cite{GFPGAN} are all proposed GAN-based methods. It can be seen that their restoration is more in line with the view of the real world in terms of the realness of the face than traditional methods, but it is not as good as the method based on diffusion models (namely DifFace and our method) in maintaining and predicting the original image information.

\subsection{Ablation Studies}\label{sec:ablation}
To better understand the roles of different components of DiffBFR, we conduct ablation studies. The first part is denoted by IRM-s, which used 1-stage DPM without introducing a cascade approach. The second part is denoted by IRM-c, which used 2-stage CDM with the traditional sampling process. The third part is denoted by IRM-c-t, which used 2-stage CDM with the Truncated Sampling Module in the second stage, that is the complete IRM in our DiffBFR. The last part is denoted by TEM-w, which added the advanced unconditional DDPM in TEM as the justify module.

We perform BFR on the CelebA-Test dataset to evaluate different components of DiffBFR. The LQ images are synthesized by the degradation model in Eq.\eqref{formula_degradation}. As shown in Figure.~\ref{AblationResultFigure}, IRM-s does not apply to the degradation model with uncertain parameters and combining multiple degradation modes, and the obtained inference image still has residual blur and noise, and the improvement of image resolution is not obvious. IRM-c decomposes the restoration process in different resolutions, and it can be clearly seen from the image that the blur degree is reduced, but there is still obvious noise residual. To remove the noise residue in the image and generate relatively detailed face information faithfully, IRM-c-t changes its sampling process. It can be clearly seen from the output of IRM-c-t that the noise added in the diffusion process is easy to be left when restoring the severely degraded image. Table.~\ref{ablationResult} lists metric results of ablation experiments. We found that after adding Truncated Sampling Module in IRM, the image noise is effectively reduced from the qualitative perspective, and FID and LPIPS are significantly reduced from the quantitative perspective. TEM-w achieves considerable results as shown in the Table.~\ref{ablationResult}, reducing indicators FID and LPIPS effectively and making the image distribution close to the real face image distribution. In Figure.~\ref{AblationResultFigure}, it is shown that this component restores local over-smoothness in details such as eyes and teeth, and the detail contour of the face is more natural and in line with the real face. Overall, DiffBFR shows superior performance to these partial components, demonstrating the efficacy of our theoretical proof.

Additionally, we assume our DiffBFR three stages respectively to explore extra parameters. 
In the sampling process of IRM which contains two stages, low-resolution in IRM(-1) and high-resolution in IRM(-2), the selection of the super-parameter $N_1$ depends on the output quality of IRM(-1) and the precision of network prediction in IRM(-2). 
The ablation results of the value of $N_1$ and $N_2$ are shown in Table.~\ref{ablationResult_about_N1} and Table.~\ref{ablationResult_about_N2}. 

\begin{table}[t!]
\setlength{\abovecaptionskip}{0.cm}
\setlength{\belowcaptionskip}{-0.cm}
    \begin{center}
    \setlength\tabcolsep{7.0pt}
    \caption{Ablation study results about $N_1$ in the sampling process of IRM(-2) on \textbf{CelebA-Test}. We choose $N_1=1000$.}
    \label{ablationResult_about_N1}
    \begin{tabular}{r|cc|ccc}
    \bottomrule
    $N_1$ & SSIM$\uparrow$ & PSNR$\uparrow$ & FID$\downarrow$ & NIQE$\downarrow$ & LPIPS$\downarrow$ \\
    \hline
     200 & 0.6759&	25.4322&	22.5438&	5.6177&	0.2759 \\
     600 & 0.6604&	25.0672&	18.5485&	5.5000&	0.2617 \\
    \textbf{1000} & 0.6494&	24.727&	19.6023&	5.4831&	0.2546 \\
     1400 & 0.6395&	24.2919&	21.4712&	5.4669&	0.2725\\
     1800 & 0.62746&	23.2766&	22.8638&	5.6274&	0.2880 \\
    \toprule
    \end{tabular}
    \end{center}
    \vspace{-4mm}
\end{table}

\begin{table}
\setlength{\abovecaptionskip}{0.cm}
\setlength{\belowcaptionskip}{-0.cm}
    \begin{center}
    \setlength\tabcolsep{8.0pt}
    \caption{Ablation study results about $N_2$ in the sampling process of TEM on \textbf{CelebA-Test}. We choose $N_2=100$.}
    \label{ablationResult_about_N2}
    \begin{tabular}{r|cc|ccc}
    \bottomrule
    $N_2$ & SSIM$\uparrow$ & PSNR$\uparrow$ & FID$\downarrow$ & NIQE$\downarrow$ & LPIPS$\downarrow$ \\
    \hline
     80 & 0.6555&	24.7671&	16.4227&	5.5868&	0.2534 \\
     \textbf{100} & 0.6553&	24.7485&	16.4902&	5.5990&	0.2535 \\
     120 & 0.6550&	24.7236&	16.5395&	5.5956&	0.2540\\
     150 & 0.6535&	24.6658&	16.8194&	5.6041&	0.2551 \\
    \toprule
    \end{tabular}
    \end{center}
    \vspace{-4mm}
\end{table}

\subsection{Discussion}
\paragraph{\textbf{Advantages}}
\textbf{(1)} Our method DiffBFR is closer to GroundTruth in the restoration effect, especially in the image color intensity and light intensity, which restores the original image to a greater extent. 
\textbf{(2)} Inference images of DiffBFR are more realistic than those of GAN-based methods. Restored images based on GAN methods pay attention to the integrity of prior knowledge, which is easy to cause huge changes to the whole facial features, while our method restores the details and retains the structural information of the original HQ image simultaneously. 
\textbf{(3)} One low-quality image can directly and reasonably correspond to several different HQ images, so the fixed mapping relationship limits the various possibilities of restoration. While DiffBFR has a certain randomness in the sampling process, which can give multiple reasonable reasoning images at the same time to deal with various possible restoration scenarios.
\paragraph{\textbf{Limitations}}
\textbf{(1)} Our method inherits the characteristics of diffusion models in the inference process, and runs for a long time. Although the Truncated Module reduces the sampling time by half, it is still longer than the running time of GAN-based methods. It needs to be further optimized for accelerated sampling in the future. 
\textbf{(2)} Compared to SR3~\cite{SR3}, a super-resolution method based on diffusion models, the parameter scale of our training model is larger, which is caused by the cascaded multi-stage model, and also for the task of image restoration with more severe degradation rather than just clean image super-resolution.
\section{Conclusion}
We have proposed DiffBFR, a face image restoration model for blind degradation based on pure diffusion models, motivated by its superiority over GANs on avoiding training collapse and generating long-tail distribution. By embedding prior into diffusion models, our model learned to generate HQ face images from randomly severely degraded ones. Specifically, we proposed two modules IRM and TEM to restore fidelity and realistic details respectively. The derivation of the theoretical boundary and the demonstration of the experimental images show the advantages of the model, and compared with recent SOTA methods, the qualitative and quantitative results are better. In the future, we will extend DiffBFR to much more severe degraded images to restore correct and realistic details. 


\newpage
\bibliographystyle{ACM-Reference-Format}
\balance
\bibliography{sample-base}

\newpage
\nobalance
\appendix

\section{Proofs}
\begin{proposition}[\textbf{IRM}]
Given a LQ image $x_0$ and HQ image $y_0$, 
we denote the evidence lower bound (ELBO) of vanilla diffusion, and diffusion with truncated sampling as $L_{DDPM}$ and $L_{IRM}$, respectively. Then, we have
    \begin{equation}
		\begin{aligned}
			L_{DDPM}\leq L_{IRM}.
		\end{aligned}
    \label{prop1}
    \end{equation}
\end{proposition}

 \noindent\textbf{Proof.} First, we can give the ELBO expression \eqref{formula3} for conditional DDPM.
    \begin{equation}
		\begin{aligned}
			\log p_{\theta}(y_0|x)&\geq E_{q(y_{1:T}|y_0)}[\log\frac{p_{\theta}(y_{0:T}|x)}{q(y_{1:T}|y_0)}]\\
			&=-D_{KL}(q(y_T|y_0)||p(y_T|x))\\&-\sum_{t=2}^{T}D_{KL}(q(y_t|y_{t-1})||p_{\theta}(y_{t-1}|y_t,x))\\&+\log p_{\theta}(y_0|y_1,x)=L_{DDPM}
		\end{aligned}
		\label{formula3}
    \end{equation}
In formula \eqref{formula3}, we focus on the first term, i.e
    \begin{equation}
		\begin{aligned}
			D_{KL}(q(y_T|y_0)||p(y_T|x)):=L_T
		\end{aligned}
    \end{equation}
In the DDPM method without Truncated Sampling Module, this term is
    \begin{equation}
		\begin{aligned}
			\rightarrow &D_{KL}(q(y_T|y_0)||p(y_T)):=L_{T|DDPM}\\
			&p(y_T)=N(y_T|0,I)
		\end{aligned}
    \end{equation}	
After changing the sampling method (IRM), this term is
    \begin{equation}
		\begin{aligned}
			\rightarrow &D_{KL}(q(y_T|y_0)||q(y_T|x)):=L_{T|IRM}\\
			&q(y_T|x)=N(y_T|\sqrt{\gamma_T}x, (1-\gamma_T)I)
		\end{aligned}
    \end{equation}
    \begin{equation}
		\begin{aligned}
			L_{T|IRM}=\frac{1}{2}\frac{\gamma_T}{1-\gamma_T}\parallel x-y_0 \parallel^2
		\end{aligned}
		\label{prop1_2}
    \end{equation}
When T cannot take positive infinity, and $x$ as the LQ image itself has partial information, we have
    \begin{equation}
		\begin{aligned}
			L_{T|DDPM} \geq L_{T|IRM}
		\end{aligned}
    \end{equation}
Then we prove the formula~\ref{prop1}.

\begin{proposition}[\textbf{TEM}]
Assume that the LQ image input is $x$, the HQ image is $y$, and the inference image is $y'$. It can be proved that the FID of the resulting image distribution after TEM is lower than that before TEM. We have
    \begin{equation}
		\begin{aligned}
			FID(x,y)>FID(y',y).
		\end{aligned}
    \end{equation} 
 \end{proposition}
  
\noindent\textbf{Proof.} Because $x=\Phi(y)$ is a pair of LQ-HQ image pairs, a sufficiently large $N$ can be satisfied by the formula~\ref{prop2_1} during the diffusion process of adding noise.
    \begin{equation}
		\begin{aligned}
			FID(x,y)>FID(x_N,y_N)
		\end{aligned}
	\label{prop2_1}
    \end{equation}
For $x_N$ and $y_N$, in the same unconditional denoise process, we could sample $x_0$ and $y_0$, respectively. In addition, since the unconditional diffusion model maps the completely Gaussian random distribution to the real distribution of data in the sampling process, namely the HQ image distribution here, we can obtain
    \begin{equation}
		\begin{aligned}
			|FID(x_0, y_0)-FID(x_N, y_N)|<|FID(x,y)-FID(x_N, y_N)|
		\end{aligned}
	\label{prop2_2}
    \end{equation}
At this time, $x_0$is inference image and $y_0$is HQ image, and the formulas~\ref{prop2_1} and~\ref{prop2_2} can be deduced
    \begin{equation}
		\begin{aligned}
			FID(x,y)>FID(x_0,y_0)=FID(y',y).
		\end{aligned}
    \end{equation} 

\section{Inference Process}
Algorithm \ref{algorithm2} describes the inference process of DiffBFR, corresponding to Figure.~\ref{framework}.

\begin{algorithm}[t]
\SetAlgoLined
\KwIn{Low-quality image $x$, prediction networks $\epsilon_{\theta_1}(x'_t, t, x_0)$, $\epsilon_{\theta_2}(y'_t, t, resize(x'_0))$ and $\epsilon_{\theta_3}(y_t, t)$, parameter $N_1, N_2$}
\KwOut{inference high quality image $y_0$}
    $x'_T\sim N(0,I)$\\
    \For{$t=T,...,1$}{
    $\epsilon\sim N(0,I)$ if $t>1$, else $\epsilon=0$\\
    $x'_{t-1}=\frac{1}{\sqrt{\alpha_t}}(x'_t-\frac{1-\alpha_t}{\sqrt{1-\gamma_t}}\epsilon_{\theta_1}(x'_t, t, x_0))+\sqrt{1-\alpha_t}\epsilon$\\
    }
    $y'_{N_1}\sim q(y'_{N_1}|y'_0=resize(x'_0))$\\
    \For{$t=N_1,...,1$}{
    $\epsilon\sim N(0,I)$ if $t>1$, else $\epsilon=0$\\
    $y'_{t-1}=\frac{1}{\sqrt{\alpha_t}}(y'_t-\frac{1-\alpha_t}{\sqrt{1-\gamma_t}}\epsilon_{\theta_2}(y'_t, t, resize(x'_0))+\sqrt{1-\alpha_t}\epsilon$\\
    }
    $y_{N_2}\sim q(y_{N_2}|y_0=y'_0)$\\
    \For{$t=N_2,...,1$}{
    $\epsilon\sim N(0,I)$ if $t>1$, else $\epsilon=0$\\
    $y_{t-1}=\frac{1}{\sqrt{\alpha_t}}(y_t-\frac{1-\alpha_t}{\sqrt{1-\gamma_t}}\epsilon_{\theta_3}(y_t, t))+\sqrt{1-\alpha_t}\epsilon$\\
    }
  \caption{Inference process}
  \label{algorithm2}
\end{algorithm}

\section{Additional Results on MNIST}
We present additional experimental details and results on the toy MNIST dataset, referring to Section 4.2, the main paper of our study. Table.~\ref{MNIST_experiment} shows parameter details of two toy models. 

\begin{table}[htbp]
\caption{Parameter details of toy models used in the main paper.}
    \begin{center}
    \begin{tabular}{c|cc}
    \bottomrule
    \textbf{Method} & \textbf{GAN-based} & \textbf{DPM-based} \\
    \hline
     Total parameters & 1.51M & 1.44M	 \\

     Total memory & 1.44Mib & 1.38Mib\\
    \toprule
    \end{tabular}
    \end{center}
        \label{MNIST_experiment}
\end{table}

\begin{table*}[htbp]
\caption{Configuration details in Section 5 of the main paper. Both Model-1 and Model-2 are trained on NVIDIA RTX 3090.}
\centering
\setlength\tabcolsep{8.0pt}
\begin{tabular}{c|c|c|c|c}
\bottomrule
\multicolumn{2}{c|}{\textbf{Details}} & \textbf{Model-1/IRM(-1)}& \textbf{Model-2/IRM(-2)} & \textbf{Model-3/TEM}\\
\hline
\multirow{7}*{diffusion model} & Input size& $128\times128$ & $512\times512$ & $512\times512$\\

~& Output size& $128\times128$ & $512\times512$ & $512\times512$\\
~& conditional & true & true & false\\
 ~& Time step& 2000& 2000& 1000\\
 ~& Beta schedule& [1e-6, 1e-2]& [1e-6, 1e-2] & [1e-4, 2e-2]\\
 ~& Loss type& $L_1$& $L_1$& $L_1$ \\
 ~& Sampler time step started& 2000 & 1000 & 100\\
 \hline
 \multirow{5}*{UNet} & channel multiplier&[1, 2, 4, 8, 8] & [1, 2, 4, 8, 8, 16, 16] & [1, 2, 4, 8, 8, 16, 16]\\
  ~& In channel& 6 &6 & 3\\
  ~& Out channel& 3 &3 & 6\\
  ~& Inner channel& 64 &64 & 32\\
  ~& Attention resolutions& [16] & [16] & [32, 16, 8]\\
\toprule
\end{tabular}
\label{experiments_setting}
\end{table*}
\section{Additional Comparison Results}
\begin{figure*}[htbp]
\begin{center}
    \includegraphics[width=0.8\linewidth]{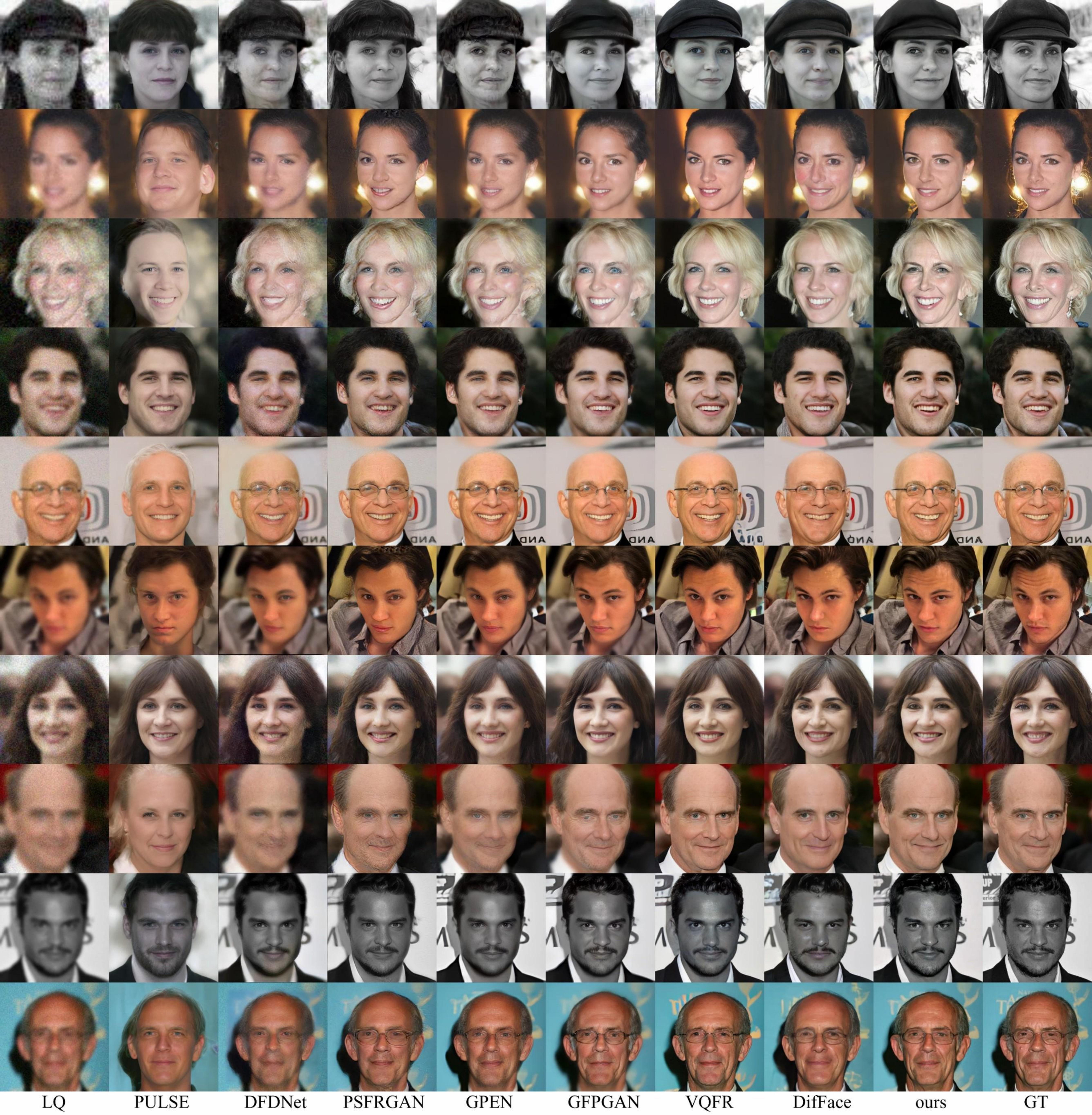}
\end{center}
    \vspace{-5mm}
   \caption{Qualitative comparisons on the \textbf{CelebA-Test} for blind face restoration. Our DiffBFR performs well in both detail complement and hue preservation. \textbf{Zoom in for the best view}.}
\label{AdditionalVisualResults}
\end{figure*}

Table.~\ref{experiments_setting} provides more details of models used in Section 5 of the main paper.
In Figure.~\ref{AdditionalVisualResults}, we present additional comparison results of DiffBFR against other state-of-the-art methods including PULSE, DFDNet, PSFRGAN, GPEN, GFPGAN, VQFR, and DifFace.

\end{document}